\pdfoutput=1

\documentclass[11pt]{article}

\usepackage[]{acl}

\usepackage{times}
\usepackage{latexsym}

\usepackage[T1]{fontenc}

\usepackage[utf8]{inputenc}

\usepackage{microtype}
\usepackage{inconsolata}
\usepackage{times}
\usepackage{pdfpages}
\usepackage{latexsym}
\usepackage{booktabs}
\usepackage{graphicx}
\usepackage{multirow}
\usepackage{times}
\usepackage{amsfonts,amsmath,amssymb,bm}
\usepackage{graphicx}
\usepackage{xcolor}
\usepackage{setspace}
\usepackage{pdflscape}
\usepackage{pdfpages}
\usepackage{enumitem}
\usepackage{rotating}
\usepackage{caption}
\usepackage{subcaption}

\newcommand{\red}[1]{\textcolor{red}{#1}}
\newcommand{\blue}[1]{\textcolor{blue}{#1}}

\usepackage[english]{babel}
\usepackage{hyperref}
\addto\captionsenglish{%
}
\addto\extrasenglish{%
}

\title{Named Entity Recognition in Twitter: \\ A Dataset and Analysis on Short-Term Temporal Shifts}

\author{
  Asahi Ushio$^1$, Leonardo Neves$^2$, Vítor Silva$^2$, Francesco Barbieri$^2$, Jose Camacho-Collados$^1$ \\
  $^1$Cardiff NLP, School of Computer Science and Informatics, Cardiff University, United Kingdom\\
  \texttt{\{UshioA,CamachoColladosJ\}@cardiff.ac.uk}\\
  $^2$Snap Inc., Santa Monica, CA, United States \\
  \texttt{\{lneves,vsilvasousa,fbarbieri\}@snap.com} \\
}

\begin{document}

\maketitle
\begin{abstract}
Recent progress in language model pre-training has led to important improvements in Named Entity Recognition (NER). Nonetheless, this progress has been mainly tested in well-formatted documents such as news, Wikipedia, or scientific articles. In social media the landscape is different, in which it adds another layer of complexity due to its noisy and dynamic nature. In this paper, we focus on NER in Twitter, one of the largest social media platforms, and construct a new NER dataset, \emph{TweetNER7}, which contains seven entity types annotated over 11,382 tweets from September 2019 to August 2021. The dataset was constructed by carefully distributing the tweets over time and taking representative trends as a basis. Along with the dataset, we provide a set of language model baselines and perform an analysis on the language model performance on the task, especially analyzing the impact of different time periods. In particular, we focus on three important temporal aspects in our analysis: short-term degradation of NER models over time, strategies to fine-tune a language model over different periods, and self-labeling as an alternative to lack of recently-labeled data. TweetNER7 is released publicly\footnote{\url{https://huggingface.co/datasets/tner/tweetner7}} along with the models fine-tuned on it\footnote{NER models have been integrated into TweetNLP \cite{camacho2022tweetnlp} and can be found at \url{https://github.com/asahi417/tner/tree/master/examples/tweetner7_paper}}.

\end{abstract}

\section{Introduction}
Named Entity Recognition (NER) is a long-standing NLP task that consists of identifying an entity in a sentence or document, and classifying it into an entity-type from a fixed typeset. One of the most common and successful types of NER system is achieved by fine-tuning pre-trained language models (LMs) on a human-annotated NER dataset with token-wise classification \cite{peters-etal-2018-deep, howard-ruder-2018-universal,GPT,GPT2,devlin-etal-2019-bert}. Remarkably, LM fine-tuning based NER models \cite{yamada-etal-2020-luke,li-etal-2020-dice} already achieve over 90\% F1 score in standard NER datasets such as CoNLL2003 \cite{tjong-kim-sang-de-meulder-2003-introduction} and OntoNotes5 \cite{hovy-etal-2006-ontonotes}.
However, NER is far from being solved, specialized domains such as financial news \cite{salinas-alvarado-etal-2015-domain}, biochemical \cite{collier-kim-2004-introduction}, or biomedical \citep{wei2015overview,li2016biocreative} still pose additional challenges \cite{ushio-camacho-collados-2021-ner}. Lower performance in these domains may be attributed to various factors such as the usage specific terminologies within those domains, which LMs have not seen while pre-training \cite{lee2020biobert}.

Among recent studies, social media has been acknowledged as one of the most challenging domains for NER \cite{derczynski-etal-2016-broad,derczynski-etal-2017-results}. Social media texts are generally more noisy and less formal than conventional written languages in addition to its vocabulary specificity.
In social media, there is another particular feature that needs to be addressed, which is the presence of (quick)  temporal shifts in the text semantics \cite{rijhwani-preotiuc-pietro-2020-temporally}, where the meaning of words is constantly changing or evolving over time. This is a general issue with language models \cite{lazaridou2021mind}, but it is especially relevant given the dynamic landscape and immediacy present in social media \cite{del-tredici-etal-2019-short}.
There have been a few specific approaches to deal with the temporal shifts in social media. For instance, \citet{loureiro-etal-2022-timelms} addressed this issue by pre-training language models on a large tweet collection from different time period, highlighting the importance of having an up-to-date language model. \citet{agarwal2021temporal} studied the temporal-shift in various NLP tasks including NER and analyzed methods to overcome the temporal-shift with strategies such as self-labeling. 

In this paper, we propose a new NER dataset for Twitter (\emph{TweetNER7} henceforth).
TweetNER7 contains tweets from diverse topics that are distributed uniformly from September 2019 to August 2021. It contains 11,382 annotated tweets in total, spanning seven entity types (\emph{person}, \emph{location}, \emph{corporation}, \emph{creative work}, \emph{group}, \emph{product}, and \emph{event}).
To the best of our knowledge, TweetNER7 is the largest Twitter NER datasets with a high coverage of entity types TTC \cite{rijhwani-preotiuc-pietro-2020-temporally} contains about same amount of annotation yet with three entity types, while WNUT17 \citep{derczynski-etal-2017-results} has six entity types yet suffer from very small annotations.
The tweets for TweetNER7 were collected by querying tweets with weekly trending keywords so that the tweet collection covers various topics within the period, and we further removed near-duplicated tweets and irrelevant tweets without any specific topics in order to improve the quality of tweets.
We provide baseline results with language model fine-tuning that showcases the difficulty of TweetNER7, especially when dealing with time shifts. 
Finally, we provide a temporal analysis with different strategies including self-labeling, which does not prove highly beneficial in our context, and provide insights in the model inner working and potential biases.

\section{Related Work}

There is a large variety of NER datasets in the literature. CoNLL2003 \cite{tjong-kim-sang-de-meulder-2003-introduction} and OntoNotes5 \cite{hovy-etal-2006-ontonotes} are widely used common NER datasets in the literature, where the texts are collected from public news, blogs, and dialogues. WikiAnn \citep{pan-etal-2017-cross} and MultiNERD \cite{tedeschi-navigli-2022-multinerd} are both multilingual NER datasets where the training set is constructed by distant-supervision on Wikipedia and BabelNet.
As far as domain-specific NER datasets are concerned, FIN \cite{salinas-alvarado-etal-2015-domain} is a NER dataset of financial news, while
BioNLP2004 \cite{collier-kim-2004-introduction} and BioCreative \cite{wei2015overview,li2016biocreative} are both constructed from scientific documents of the biochemical and biomedical domains. However, none of these datasets address the same challenges posed by the social media domain.

In the social media domain, the pioneering Broad Twitter Corpus (BTC) NER dataset \cite{derczynski-etal-2016-broad} included users with  different demographics with the aim to investigate spatial and temporal shift of semantics in NER. More recently, the test set of WNUT2017 \cite{derczynski-etal-2017-results} contained unseen entities in the training set from broader social media including Twitter, Reddit, YouTube, and StackExchange.
The recent TweeBankNER dataset \cite{jiang-EtAl:2022:LREC2} annotated TweeBank \cite{liu-etal-2018-parsing} with entity labels to investigate the interaction between syntax and NER.

The most similar dataset to ours is the Temporal Twitter Corpus (TTC) NER dataset. \cite{rijhwani-preotiuc-pietro-2020-temporally}, which was also aimed at analysing the temporal effects of NER in social media. For this dataset, 2,000 tweets every year from 2014 to 2019 were annotated. 
In general, however, these social media datasets suffer from limited data, non-uniform distribution over time, or limited entity types (see Subsection \ref{sec:statistics} for more details). In this paper, we contribute with a new NER dataset (TweetNER7) based on recent data until 2021, which is specifically designed to analyze temporal shifts in social media.

\section{TweetNER7: Dataset Construction, Statistics and Baselines}
In this section, we present our time-aware NER dataset from publicly available tweets with seven general entity types, which we refer as \emph{TweetNER7}. In the following subsections, we describe the data collection (Subsection \ref{sec:dataset-collection}) and annotation (Subsection \ref{sec:dataset-annotation}) processes. We also share relevant statistics (Subsection \ref{sec:statistics}) and baseline results (Subsection \ref{sec:baseline}) of our dataset.

\subsection{Data Collection} \label{sec:dataset-collection}

This NER dataset annotates a similar tweet collection used to construct TweetTopic \cite{dimosthenis-etal-2022-twitter}. The main data consists of tweets from September 2019 to August 2021 with roughly same amount of tweets in each month. This collection period makes it suitable for our purpose of evaluating short-term temporal-shift of NER on Twitter. 
The original tweets were filtered by leveraging weekly trending topics as well as by various other types of filtering see \newcite{dimosthenis-etal-2022-twitter} for more details on the collection and filtering process).
The collected tweets were then split into two periods: September 2019 to August 2020 (2020-set) and September 2020 to August 2021 (2021-set).

\subsection{Dataset Annotation} \label{sec:dataset-annotation}

\begin{figure*}[t]
    \centering
    \includegraphics[width=2.0\columnwidth]{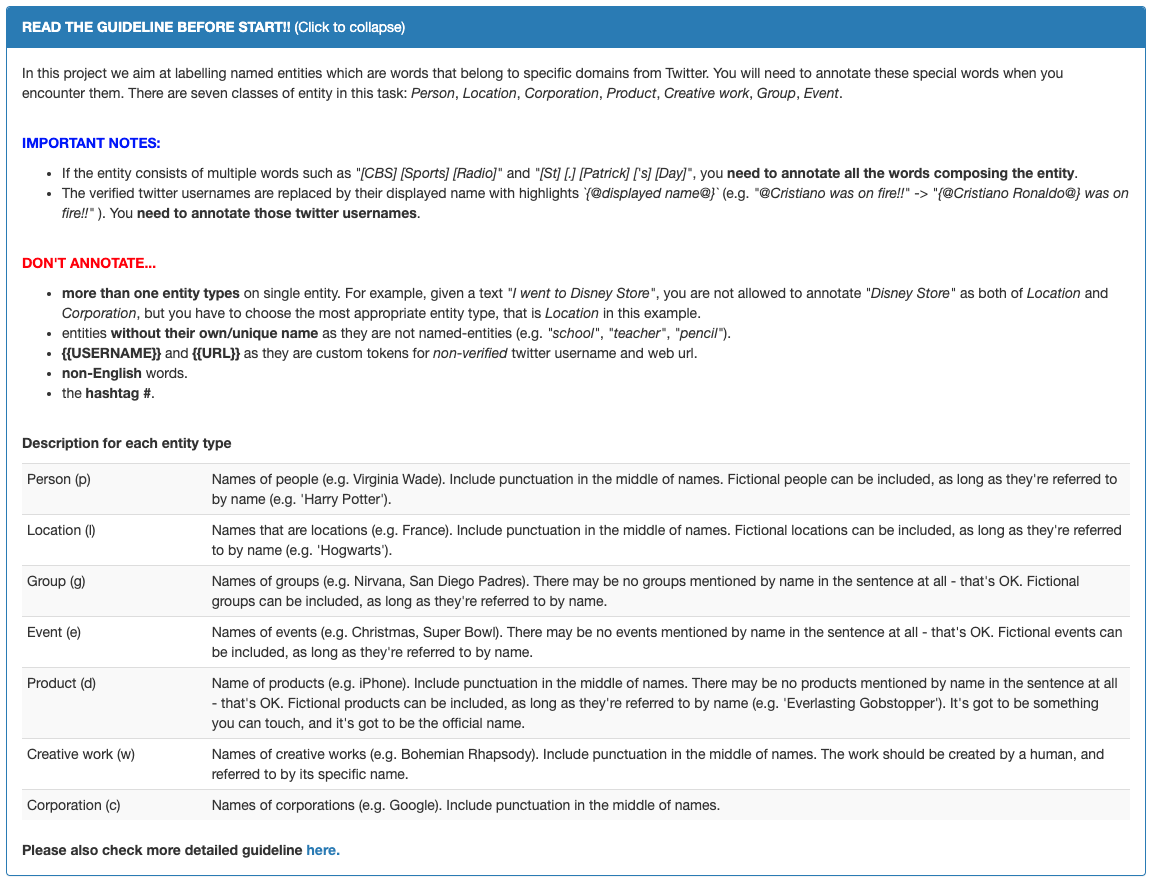}
    \caption{
    The instructions shown to the annotators during the annotation phase.
    }
    \label{fig:instruction}
\end{figure*}

\noindent \textbf{Annotation.} To attain named-entity annotations over the tweets, we conducted a manual annotation on Amazon Mechanical Turk with the interface shown in \autoref{fig:instruction}.
We split tweets into two periods:
September 2019 to August 2020 (2020-set) and 
September 2020 to August 2021 (2021-set), and randomly sampled 6,000 tweets from each period, which were annotated by three annotators, collecting 36,000 annotations in total.
As the entity types, we employed seven labels: \emph{person}, \emph{location}, \emph{corporation}, \emph{creative work}, \emph{group}, \emph{product}, and \emph{event}. We followed \newcite{derczynski-etal-2017-results} for the selection of the first six labels, and additionally included \emph{event}, as we found a large amount of entities for events in our collected tweets. 

\noindent \textbf{Pre-processing.} 
We pre-process tweets before the annotation to normalize some artifacts, converting URLs into a special token $\{\{\texttt{URL}\}\}$ and non-verified usernames into $\{\{\texttt{USERNAME}\}\}$.
For verified usernames, we replace its display name with symbols $\texttt{@}$.
For example, a tweet
\begin{quote}
\small
\begin{verbatim}
Get the all-analog Classic Vinyl Edition
of "Takin' Off" Album from @herbiehancock
via @bluenoterecords link below: 
http://bluenote.lnk.to/AlbumOfTheWeek
\end{verbatim}
\end{quote}
is transformed into the following text.
\begin{quote}
\small
\begin{verbatim}
Get the all-analog Classic Vinyl Edition
of "Takin' Off" Album from {@Herbie Hancock@}
via {{USERNAME}} link below: {{URL}}
\end{verbatim}
\end{quote}
We ask annotators to ignore those special tokens but label the verified users' mentions.

\noindent \textbf{Quality Control.}
Since we have three annotations per tweet, we control the quality of the annotation by taking the agreement into account.
We disregard the annotation if the agreement is 1/3, and manually validate the annotation if it is 2/3, which happens for roughly half of the instances.

\subsection{Statistics}
\label{sec:statistics}
This subsection provides an statistical analysis of (i) our dataset, (ii) our dataset in comparison with other Twitter NER datasets, and (iii) our dataset distribution over time.

\noindent\textbf{Statistics of TweetNER7.}
TweetNER7 contains 5,768 and 5,614 tweets annotated in each period of 2020 and 2021, which are then split into training / validation / test sets for each year.
Since the 2020-set is for model development, we consider 80\% of the dataset as training set and 10\% for validation and test sets.
Meanwhile, the 2021-set is mainly devised for model evaluation to measure the temporal adaptability, so we take the majority of the 2021-set (50\%) as the test set and split the rest into training and validation set with the same ratio of training and validation set of the 2020-set.
Table~\ref{tab:data-stats} summarizes the number of the entities as well as the instances in each subset of TweetNER7. We can observe a large gap between frequent entity types such as \emph{person} and rare entity types as \emph{location}, while the distribution of the entities are roughly balanced across subsets.
We also report entity diversity, which we define as the percentage of unique entities with respect to the total number of entities. Entity types such as \emph{product} contain a relatively large number of duplicates (ranging between 56.2\% and 76.4\% entity diversity scores), while other types such as creative work are more diverse (ranging between 80.1\% and 93.2\%).

\begin{table}[t]
\centering
\scalebox{0.78}{
\begin{tabular}{@{}l@{\hspace{5pt}}r@{\hspace{5pt}}r@{\hspace{5pt}}r@{\hspace{5pt}}r@{\hspace{5pt}}r@{\hspace{5pt}}r@{}}
\toprule
Period          & \multicolumn{3}{c}{{2020-set}} & \multicolumn{3}{c}{{2021-set}} \\
Split           & Train    & Valid & Test & Train & Valid & Test \\
\midrule
\multicolumn{7}{@{}l}{Number of Entities} \\
- corporation   & 1,700       & 203   & 191  & 902      & 102 & 900          \\
- creative work & 1,661       & 208   & 179  & 690      & 74 & 731          \\
- event         & 2,242       & 256   & 265  & 968      & 131 & 1,097        \\
- group         & 2,242       & 227   & 311  & 1,313    & 227 & 1,516        \\
- location      & 1,259       & 181   & 165  & 697      & 72 & 716          \\
- person        & 4,666       & 598   & 596  & 2,362    & 283 & 2,712        \\
- product       & 1,850       & 241   & 220  & 926      & 111 & 972          \\
All           & 15,620      & 1,914 & 1,927& 8,864      & 1,000 & 8,644      \\\midrule
\multicolumn{7}{@{}l}{Entity Diversity} \\
- corporation   & 69.9 & 92.6 & 90.1 & 72.1 & 85.3 & 74.3 \\
- creative work & 80.1 & 92.8 & 91.6 & 89.0 & 93.2 & 91.0 \\
- event         & 71.1 & 90.6 & 84.2 & 75.9 & 89.3 & 70.9 \\
- group         & 66.7 & 86.8 & 81.7 & 66.0 & 86.3 & 66.2 \\
- location      & 66.4 & 80.7 & 81.2 & 67.9 & 88.9 & 64.9 \\
- person        & 68.4 & 85.6 & 83.6 & 77.3 & 90.1 & 77.7 \\
- product       & 56.2 & 71.4 & 76.4 & 60.3 & 79.3 & 56.6 \\\midrule
Number of Tweets& 4,616        & 576   & 576  & 2,495     & 310   & 2,807 \\ \bottomrule
\end{tabular}
}
\caption{
Number of entities, tweets, and entity diversity in each data split and period, where the 2020-set is from September 2019 to August 2020, while the 2021-set is from September 2020 to August 2021.
}
\label{tab:data-stats}
\end{table}

\noindent\textbf{Comparison with other Twitter NER Datasets.} In \autoref{tab:stats-comparison}, we compare TweetNER7 against existing NER datasets for Twitter, which highlights the large number of annotations of TweetNER7 for our covered period. 
TweetNER7 and TTC are the overall largest datasets with more than 10k annotations, but TTC covers only three entities, which may be insufficient for certain practical use cases given the diversity of text in social media context \cite{derczynski-etal-2017-results}. In contrast, TweetNER7 has the highest coverage of entity types among all NER datasets in Twitter, including all the entity types from existing datasets. 
In addition to the large amount of annotations and a high coverage of entity types, TweetNER7 includes recent tweets from 2019 to 2021, from which most corpus used in pre-training language models do not contain any text \cite{devlin-etal-2019-bert,RoBERTa,nguyen-etal-2020-bertweet}.
Assuming we tackle NER by language model fine-tuning, this fact makes the task further challenging, since language models have never seen the emerging entities from the period during its pre-training phase.


\begin{table}[t]
\centering
\scalebox{0.75}{
\begin{tabular}{@{\hspace{5pt}}l@{\hspace{5pt}}r@{\hspace{5pt}}r@{\hspace{5pt}}r@{\hspace{5pt}}r@{\hspace{5pt}}}
\toprule
Dataset     & Annotations   & Entities    & Domain        & Year      \\\midrule
BTC         & 9,339         & 3         & Twitter       & 2009-2015 \\
WNUT2017   & 5,690         & 6         & Twitter+  & 2010-2017 \\ 
TTC         & 11,969        & 3         & Twitter       & 2014-2019 \\
TweeBankNER & 3,547         & 4         & Twitter       & 2016      \\
\midrule
TweetNER7   & 11,382        & 7         & Twitter       & 2019-2021 \\ \bottomrule
\end{tabular}
}
\caption{Number of annotated instances in TweetNER7 and comparison NER datasets for Twitter.}
\label{tab:stats-comparison}
\end{table}

\begin{table}[t]
\centering
\scalebox{0.78}{
\begin{tabular}{@{}l@{\hspace{5pt}}r@{\hspace{5pt}}r@{\hspace{5pt}}r@{\hspace{5pt}}r@{\hspace{5pt}}r@{\hspace{5pt}}r@{}}
\toprule
         & Jan   & Feb   & Mar   & Apr   & May   & Jun   \\ \midrule
BTC      & 2,308 & 68    & 502   & 862   & 1,074 & 1,056 \\
TTC      & 945   & 1,014 & 1,307 & 1,089 & 764   & 694   \\
TweetNER7& 957   & 943   & 939   & 937   & 951   & 931   \\\bottomrule \toprule
         & Jul   & Aug   & Sep   & Oct   & Nov   & Dec   \\\midrule
BTC      & 1,321 & 850   & 342   & 419   & 23    & 21    \\
TTC      & 760   & 754   & 889   & 958   & 958   & 866   \\
TweetNER7& 924   & 928   & 956   & 968   & 975   & 973   \\\bottomrule
\end{tabular}
}
\caption{
The number of tweets in each month from BTC, TTC, and our TweetNER7 (the counts are cumulated across years).
The normalized standard deviation across month is 7.5\% (BTC), 1.6\% (TTC), and 0.2\% (TweetNER7).}
\label{tab:monthly}
\end{table}

\noindent\textbf{Distribution over Time.} One of the TweetNER7's focus is the temporal shift in Twitter similar to BTC and TTC datasets. Retaining uniform distribution over time is essential for temporal analysis, since the amount of training instances should have an effect to the metric if it is not uniform.
\autoref{tab:monthly} shows the distribution of the instances across each month and we can confirm that TweetNER7 has a very similar amount of tweets each month, while BTC and TTC have higher variation than TweetNER7.
Moreover, \autoref{tab:yearly} compares the number of instances per year for each dataset. TweetNER7 has a an uneven distribution here due to the the selected range for each period (i.e., September 2019 to August 2021), which results in more tweets in 2020 than 2019 and 2021.

\begin{table}[t]
\centering
\scalebox{0.75}{
\begin{tabular}{@{}l@{\hspace{5pt}}r@{\hspace{5pt}}r@{\hspace{5pt}}r@{\hspace{5pt}}r@{\hspace{5pt}}r@{\hspace{5pt}}r@{\hspace{5pt}}r@{}}
\toprule
          & 2009 & 2010 & 2011 & 2012  & 2013 & 2014  & 2015  \\\midrule
BTC       & 3    & 5    & 127  & 2,414 & 275  & 6,022 & 0     \\
TTC       & 0    & 0    & 0    & 0     & 0    & 2,000 & 2,000 \\
TweetNER7 & 0    & 0    & 0    & 0     & 0    & 0     & 0     \\\bottomrule
\toprule
          & 2016  & 2017  & 2018  & 2019  & 2020  & 2021 &  \\\midrule
BTC       & 0     & 0     & 0     & 0     & 0     & 0    & \\
TTC       & 2,000 & 2,000 & 2,000 & 2,000 & 0     & 0    & \\
TweetNER7 & 0     & 0     & 0     & 1,936 & 5,768 & 3,678&\\\bottomrule
\end{tabular}
}
\caption{The number of tweets in each year from BTC, TTC, and our TweetNER7 dataset.}
\label{tab:yearly}
\end{table}

\subsection{Baseline Results}
\label{sec:baseline}
Finally, we introduce a couple of baselines with language model fine-tuning on the TweetNER7 in temporal-shift setup, where we develop models with the training and the validation set from the 2020-set, and evaluate the models on the test set of the 2021-set. 
In this setup, models are required to generalize to the text from newer period, which the model has not seen in the fine-tuning phase.

\noindent\textbf{Experimental Setting.}
We consider masked language model fine-tuning with the following LMs: BERT \cite{devlin-etal-2019-bert} and RoBERTa \cite{RoBERTa} as general-purpose LMs, and BERTweet \cite{nguyen-etal-2020-bertweet}, and TimeLMs \cite{loureiro-etal-2022-timelms} as Twitter-specific LMs. TimeLMs are based on a RoBERTa\textsubscript{BASE} architecture pre-trained on tweets collected continuously until different years: 2019, 2020, and 2021.
Model weights are taken from HuggingFace \cite{wolf-etal-2020-transformers}.\footnote{
We use \texttt{bert-base-cased} and \texttt{bert-large-cased} for BERT,
\texttt{roberta-base} and \texttt{roberta-large} for RoBERTa,
\texttt{vinai/bertweet-base} and \texttt{vinai/bertweet-large} for BERTweet, and
\texttt{cardiffnlp/twitter-roberta-base-2019-90m}, \texttt{cardiffnlp/twitter-roberta-base-dec2020}, and \texttt{cardiffnlp/twitter-roberta-base-dec2021} for TimeLMs.
}
As evaluation metrics, we consider micro/macro F1 score and type-ignored F1 score \cite{ushio-camacho-collados-2021-ner}, in which the entity type of the prediction is not considered in the evaluation (i.e., this metric only assesses whether the predicted entity is an entity or not).
The F1 scores measure the NER systems' entire performance, while the type-ignored F1 score measures the ability of identifying whether a span of text is an entity or not.
LM fine-tuning on NER relies on the T-NER library \cite{ushio-camacho-collados-2021-ner} and to find the best combination of hyper-parameters to fine-tune LMs on NER, we run two-phase grid search.
First, we fine-tune a model on every possible configuration from the search space for 10 epochs. The top-5 models in terms of micro F1 score on the validation set are selected to continue fine-tuning until their performance plateaus, and then the model that achieves the highest micro F1 score on the validation set is employed as the final model.
The search space contains 24 configurations, which consist of the following variations: learning rates from $[0.000001, 0.00001, 0.0001]$;
ratio of total training step for linear warm up of learning rate from $[0.15, 0.3]$;
whether to normalize the gradient norm or not;
and whether to add conditional random field (CRF) on top of the output logit of LM.\footnote{Other parameters are fixed: random seed is $0$ and batch size is $32$.}

\begin{table}[t]
\centering
\scalebox{0.75}{
\begin{tabular}{@{}l@{\hspace{10pt}}c@{\hspace{10pt}}c@{\hspace{10pt}}c@{}}
\toprule
\multirow{2}{*}{Model}          & Micro F1                  & Macro F1                  & Type-ig. F1 \\
                                & 2021 / 2020               & 2021 / 2020               & 2021 / 2020 \\\midrule
BERT\textsubscript{BASE}        & 60.1 / 60.9               & 54.7 / 56.5               & 75.6 / 72.4\\
BERT\textsubscript{LARGE}       & 61.4 / 62.2               & 56.1 / 58.1               & 75.9 / 73.8\\
BERTweet\textsubscript{BASE}    & 64.1 / \underline{66.4}   & 59.4 / 62.4               & 77.9 / \underline{77.7}\\
BERTweet\textsubscript{LARGE}   & 64.0 / 65.9               & 59.5 / \underline{62.6}	& 78.3 / 77.4\\
RoBERTa\textsubscript{BASE}     & 64.2 / 64.2		        & 59.1 / 60.2               & 77.9 / 74.8\\
RoBERTa\textsubscript{LARGE}    & \textbf{64.8} / 65.7      & \textbf{60.0} / 61.9	    & \textbf{78.4} / 76.1\\
TimeLM\textsubscript{2019}      & 64.3 / 65.4	            & 59.3 / 61.1	            & 77.9 / 76.6\\
TimeLM\textsubscript{2020}      & 62.9 / 64.4               & 58.3 / 60.3	            & 76.5 / 75.7\\
TimeLM\textsubscript{2021}      & 64.2 / 65.4	            & 59.5 / 61.1	            & 77.4 / 76.4 \\
\bottomrule
\end{tabular}
}
\caption{Result of temporal-shift NER on TweetNER7 where micro and macro F1 score as well as type-ignored F1 score on the test set of the 2021-set / 2020-set are reported.
The best results in each of the 2021-set / 2020-set are highlighted in bold character / underline in each metric.}
\label{tab:baseline-2020}
\end{table}

\noindent\textbf{Results.}
We report the NER results on TweetNER7 in \autoref{tab:baseline-2020}, where RoBERTa\textsubscript{LARGE} is the best across metrics. We should note, however, that the overall metrics (micro F1 lower than 65\% in all cases on the 2021 test set) are lower than those in standard NER datasets \cite{ushio-camacho-collados-2021-ner}, which highlights the difficulty of the social media and temporal-shift components in TweetNER7.
RoBERTa is also the best model among the \textsubscript{BASE} models but interestingly the TimeLM\textsubscript{2020} performs worse than other RoBERTa models.
This can be explained by the fact that TimeLM\textsubscript{2020} was pre-trained over tweets until the end of 2020. This may have let the model to over-fit to the training corpus and makes it hard to generalize on the newer test set. 
Instead, TimeLM\textsubscript{2021} shows a better performance.
\autoref{tab:baseline-2021} also reports the metrics on the 2020 test set for completeness. While that is not our primary aim, we can find an interesting result which is the superior performance of BERTweet in this case. 
This implies that a model that performs well in the same period of the training set does not guarantee an equally strong performance on an unseen period.

\begin{figure}[t]
    \centering
    \includegraphics[width=1.0\columnwidth]{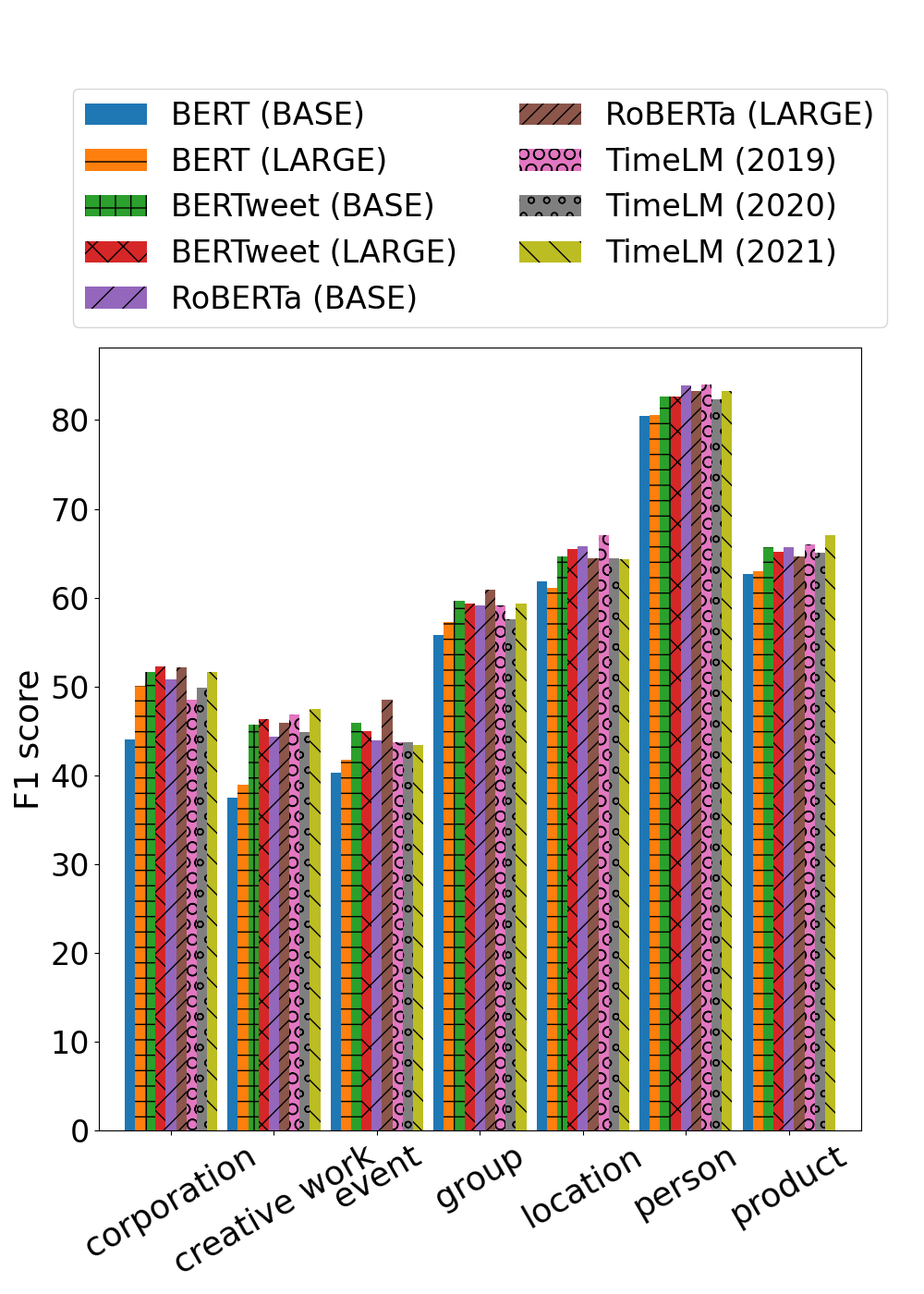}
    \caption{Entity-wise F1 score breakdown from the baseline results in the 2021 test set (\autoref{tab:baseline-2020}).
    }
    \label{fig:entity-breakdown}
\end{figure}

\noindent\textbf{Breakdown by entity type.} \autoref{fig:entity-breakdown} shows a comparison of entity-wise F1 scores over the language models, and we can see an important gap across entity types.
According to \autoref{tab:data-stats}, \emph{person} is the most frequent entity type and its F1 score is equally high (around 80\%), while \emph{creative work} and \emph{location} are the rarest entity types and hence their F1 scores are relatively low (around 40\% for \emph{creative work} and 60\% for \emph{location}). The reason why the performance for \emph{location} is better than for \emph{creative work} may be attributable to their differences in entity diversity.
As we could see from \autoref{tab:data-stats}, \emph{creative work}'s diversity is higher than \emph{location}, which means \emph{creative work} contains more variation of entities than \emph{location} while having the same amount of entities in both types, which entails a higher degree of difficulty.
This seems a consistent trend that lower entity diversity results in lower F1 score as can be seen for \emph{event} and \emph{corporation} as well, which also have a low entity diversity score.
To overcome such entity imbalance, strategies such as balancing the instances of each class could be explored \cite{li-etal-2020-dice}.

\section{Temporal Analysis}
\label{sec:temporal-shift-analysis}
To better understand the effect of the temporal-shift, we conduct three additional comparative experiments: (i) temporal vs. random splits, (ii) joint vs. continuous fine-tuning, and (iii) self-labeling as a solution to deal with temporal shifts.

\subsection{Short-Term Temporal Effect}
\label{sec:random-split} If TweetNER7 does not suffer temporal-shift, how is the model performance changed? This is a question we aim to answer in this analysis, and we create new training and validation split without temporal-shift for this purpose.
Concretely, temporal-shift usually occurs in a situation where the training and the validation sets do not contain any texts from the test period, so we keep the amount of the training/validation split as the same in Subsection \ref{sec:baseline}, but randomly sample from the full period of September 2019 to August 2021 instead of the first half period instead.
Note that we do not change the test set and make sure that each month has roughly the same amount of instances at the sampling of the new training/validation sets, to make it fair comparison with the temporal-shift result in Subsection \ref{sec:baseline}.

\autoref{tab:random-split} shows the variations of results between the random and temporal splits. As expected, the F1 scores on the 2021 test set are generally improved across all LMs, while the F1 scores on the 2020 test set are decreased.
The increase of accuracy in 2021 is achieved with the inclusion of training/validation set from 2021, and the decrease of accuracy in 2020 is caused by the reduced number of the training/validation set from the same 2020 period.
This result further highlights the benefit of having a human annotated training set from the test period, even if the time period differs in a year only. Interestingly, the results for the time-specific pre-trained TimeLMs models differ across years. Since in this paper we did not focus on the analysis of the pre-training corpora, we leave further analysis about this result for future exploration.

\begin{table}[t]
\centering
\scalebox{0.75}{
\begin{tabular}{@{}l@{\hspace{5pt}}c@{\hspace{5pt}}c@{\hspace{5pt}}c@{\hspace{5pt}}c@{\hspace{5pt}}c@{\hspace{5pt}}c@{}}
\toprule
\multirow{2}{*}{Model}  & \multicolumn{3}{c}{{2021-set}} & \multicolumn{3}{c}{{2020-set}}  \\
& Mi. F1 & Ma. F1  & T-i. F1 & Mi. F1 & Ma. F1 & T-i. F1\\
\midrule
BERT\textsubscript{BASE}    
& \blue{+0.8} & \blue{+1.2}& \blue{+0.1}       
& \blue{+0.1}& \blue{+0.3}& \blue{+0.3}       \\
BERT\textsubscript{LARGE}   
& \blue{+1.0}& \blue{+1.4}& \blue{+0.6}
& \red{-0.7} & \red{-1.0} & \red{-0.5}      \\
BERTweet\textsubscript{BASE}    
& \blue{+1.5}& \blue{+0.2}& \red{-0.1}
& \red{-2.5} & \red{-3.8} & \red{-3.3}     \\
BERTweet\textsubscript{LARGE}   
& \blue{+0.9}& \blue{+1.0}& \blue{+0.1}
& \blue{+0.1}& \blue{+0.1}& \red{-0.2}      \\
RoBERTa\textsubscript{BASE} 
& \red{-0.2} & \blue{+0.1}& \blue{+0.1}
& \red{-0.1} & \red{-0.4} & \red{-0.5}      \\
RoBERTa\textsubscript{LARGE}
& \blue{+1.5}& \blue{+1.0}& \blue{+0.6}
& \red{-1.3} & \red{-1.8} & \red{-0.6}      \\
TimeLM\textsubscript{2019}  
& \red{-1.0} & \red{-0.8} & \red{-0.5}
& \red{-1.1} & \red{-0.4} & \red{-0.4}     \\
TimeLM\textsubscript{2020}  
& \blue{+1.8}& \blue{+1.7}& \blue{+1.8}
& \blue{+0.3}& \blue{+0.2}& \blue{+0.2}       \\
TimeLM\textsubscript{2021}  
& \red{-1.0} & \red{-1.1} & \red{-0.4}
& \red{-1.7} & \red{-1.3} & \red{-1.0}
\\\bottomrule
\end{tabular}
}
\caption{
Absolute performance improvement when evaluating on the random split result over the original temporal split reported in \autoref{tab:baseline-2020}. Positive improvements are in blue and negative drops are in red.}
\label{tab:random-split}
\end{table}

\begin{table}[h]
\centering
\scalebox{0.75}{
\begin{tabular}{@{}l@{\hspace{5pt}}l@{\hspace{5pt}}c@{\hspace{5pt}}c@{\hspace{5pt}}c@{\hspace{5pt}}c@{}}
\toprule
& & Dataset      & Micro F1       & Macro F1     & Type-ig. F1        \\ \midrule
\multirow{8}{*}{\rotatebox{90}{BERT}} & 
\multirow{4}{*}{\rotatebox{90}{\textsubscript{BASE}}}      
& 2020                      & 60.1 / 60.9 & 54.7 / 56.5 & 75.6 / 72.4 \\
&& 2021                     & 60.7 / 58.4 & 55.5 / 54.2 & 75.7 / 70.9 \\
&& 2020 + 2021              & \textbf{62.3} / \underline{62.1} & \textbf{57.6} / \underline{57.7} & \textbf{76.6} / \underline{73.0} \\
&& 2020 $\rightarrow$ 2021  & 61.8 / 61.4 & 56.8 / 57.1 & 76.5 / 72.5 \\ \cline{2-6} 
& \multirow{4}{*}{\rotatebox{90}{\textsubscript{LARGE}}}     
& 2020                      & 61.4 / 62.2 & 56.1 / 58.1 & 75.9 / \underline{73.8} \\
&& 2021                     & 59.7 / 56.6 & 53.9 / 51.0 & 75.0 / 70.7 \\
&& 2020 + 2021              & \textbf{63.6} / \underline{62.5} & \textbf{59.0} / \underline{58.6} & \textbf{77.2} / 73.6 \\
&& 2020 $\rightarrow$ 2021  & 63.2 / \underline{62.5} & 57.7 / 57.9 & 76.0 / 72.5 \\\midrule
\multirow{8}{*}{\rotatebox{90}{BERTweet}} & 
\multirow{4}{*}{\rotatebox{90}{\textsubscript{BASE}}}
& 2020                      & 64.1 / \underline{66.4} & 59.4 / \underline{62.4} & 77.9 / \underline{77.7} \\
&& 2021                     & 63.1 / 62.1 & 57.4 / 57.2 & 77.9 / 76.0 \\
&& 2020 + 2021              & 65.4 / 65.7 & 60.5 / 61.6 & 79.0 / 76.9 \\
&& 2020 $\rightarrow$ 2021  & \textbf{65.8} / 65.2 & \textbf{61.0} / 61.4 & \textbf{79.1} / 76.8 \\\cline{2-6} 
& \multirow{4}{*}{\rotatebox{90}{\textsubscript{LARGE}}} 
& 2020                      & 64.0 / 65.9 & 59.5 / 62.6 & 78.3 / 77.4 \\
&& 2021                     & 62.9 / 61.6 & 58.1 / 56.8 & 76.5 / 74.5 \\
&& 2020 + 2021              & \textbf{66.5} / \underline{66.8} & \textbf{61.9} / \underline{63.1} & \textbf{79.5} / \underline{77.6} \\
&& 2020 $\rightarrow$ 2021  & 66.4 / 65.9 & 61.7 / 61.8 & 79.0 / 76.4 \\\midrule
\multirow{8}{*}{\rotatebox{90}{RoBERTa}} & 
\multirow{4}{*}{\rotatebox{90}{\textsubscript{BASE}}}   
& 2020                      & 64.2 / 64.2 & 59.1 / 60.2 & 77.9 / 74.8 \\
&& 2021                     & 61.8 / 60.5 & 57.0 / 56.1 & 76.9 / 73.8 \\
&& 2020 + 2021              & 65.2 / \underline{65.3} & \textbf{60.8} / \underline{61.7} & \textbf{78.9} / \underline{75.2} \\
&& 2020 $\rightarrow$ 2021  & \textbf{65.5} / 65.1 & 60.0 / 60.8 & 78.1 / 75.0 \\\cline{2-6} 
& \multirow{4}{*}{\rotatebox{90}{\textsubscript{LARGE}}}  
& 2020                      & 64.8 / 65.7 & 60.0 / 61.9 & 78.4 / 76.1 \\
&& 2021                     & 64.0 / 63.4 & 59.1 / 59.1 & 77.7 / 74.4 \\
&& 2020 + 2021              & 65.7 / \underline{66.3} & \textbf{61.2} / \underline{63.0} & 78.8 / \underline{76.4} \\
&& 2020 $\rightarrow$ 2021  & \textbf{66.0} / \underline{66.3} & 60.9 / 62.4 & \textbf{79.1} / \underline{76.4} \\\midrule
\multirow{12}{*}{\rotatebox{90}{TimeLM}} 
& \multirow{4}{*}{\rotatebox{90}{\textsubscript{2019}}}    
& 2020                      & 64.3 / 65.4 & 59.3 / 61.1 & 77.9 / \underline{76.6} \\
&& 2021                     & 63.2 / 61.9 & 56.7 / 56.1 & 75.7 / 73.0 \\
&& 2020 + 2021              & 65.7 / \underline{65.5} & 61.0 / \underline{61.2} & \textbf{78.9} / 76.4 \\
&& 2020 $\rightarrow$ 2021  & \textbf{65.9} / 64.8 & \textbf{61.1} / 60.6 & 78.4 / 75.5 \\\cline{2-6} 
& \multirow{4}{*}{\rotatebox{90}{\textsubscript{2020}}}    
& 2020                      & 62.9 / 64.4 & 58.3 / 60.3 & 76.5 / 75.7 \\
&& 2021                     & 64.0 / 63.1 & 58.9 / 58.5 & 77.9 / 75.3 \\
&& 2020 + 2021              & 65.3 / \underline{65.4} & \textbf{60.7} / \underline{61.4} & \textbf{78.7} / \underline{75.9} \\
&& 2020 $\rightarrow$ 2021  & \textbf{65.5} / 65.3 & 60.6 / 61.3 & 78.0 / \underline{75.9} \\\cline{2-6} 
& \multirow{4}{*}{\rotatebox{90}{\textsubscript{2021}}}    
& 2020                      & 64.2 / 65.4 & 59.5 / 61.1 & 77.4 / 76.4 \\
&& 2021                     & 63.5 / 62.3 & 58.7 / 57.9 & 77.5 / 74.1 \\
&& 2020 + 2021              & 64.5 / \underline{65.8} & 59.8 / \underline{61.9} & 77.9 / \underline{76.5} \\
&& 2020 $\rightarrow$ 2021  & \textbf{65.1} / 64.9 & \textbf{60.0} / 60.7 & \textbf{78.1} / 75.8 \\\bottomrule
\end{tabular}
}
\caption{
Results of different strategies to ingest the training set of the 2021-set in TweetNER7 for different language models ($\rightarrow$: continuous fine-tuning; +: concatenation of datasets).
The best results in each model of the 2021-set / 2020-set are highlighted in bold character / underline in each metric.}
\label{tab:baseline-2021}
\end{table}

\subsection{Continuous vs. Joint Fine-Tuning}
\label{sec:continuous-fine-tuning}

In the previous experiments we have shown the differences between training and testing on the time period or not. Instead, this analysis comes under the assumption that a labeled 2021 training set is available. Thus, the main aim of this analysis is to explore different strategies to improve the original model.
In addition to fine-tuning LMs on the combined set of the 2020-set and 2021-set as in Subsection \ref{sec:random-split}, we employed a continuous fine-tuning scheme, where we first fine-tune LMs on the 2020-set and then continue fine-tuning on the 2021-set.
\autoref{tab:baseline-2021} shows the results of all strategies for different language models.
As can be observed, 
continuous fine-tuning provides the best results in terms of micro F1 and type-ignored F1 in the 2021 test sets in most cases, although the differences with respect to the concatenation of sets are not substantial.

\subsection{Self-Labeling}
\label{sec:self-labeling}

In both Subsections \ref{sec:random-split} and \ref{sec:continuous-fine-tuning}, we compared different strategies when a human-annotated training dataset from the test period was considered, namely the training and the validation sets from the 2021-set. This shows that improvements can be obtained when the time between training and test data is reduced.
However, in many cases and real-world applications this is not practical as it requires a large amount of human resources to annotate newer tweets whenever.
Thus, we consider an alternative approach to rely on distantly annotated tweets by the already fine-tuned model. This solution was explored by \newcite{agarwal2021temporal} in a similar setting, with promising results. In this paper, we reproduced their experiments in our TweetNER7 dataset focusing on short-term temporal shift.

\begin{table}[t]
\centering
\scalebox{0.78}{
\begin{tabular}{@{}l@{\hspace{5pt}}c@{\hspace{5pt}}c@{\hspace{5pt}}c@{}}
\toprule
\multirow{2}{*}{Training Set}   & Micro F1                  & Macro F1                  & Type-ig. F1 \\
                                        & 2021 / 2020               & 2021 / 2020               & 2021 / 2020 \\\midrule
                                        
                                        2020         & \textbf{64.8} / \underline{65.7} & \textbf{60.0} / \underline{61.9} & 78.4 / 76.1 \\\midrule
2020-extra                                    
& 64.6 / 65.5               & 59.3 / 61.4               & 78.6 / 76.2 \\
2020 + 2020-extra                            
& 64.7 / 65.2               & 59.6 / 61.0               & \textbf{78.7} / 76.8 \\
2020 $\rightarrow$ 2020-extra                       
& 64.6 / 65.5               & 59.5 / 61.5               & 78.6 / 76.4 \\
2021-extra                                    
& 64.2 / \underline{65.7}               & 59.3 / 61.8               & 78.2 / \underline{76.9} \\
2020 + 2021-extra                            
& 64.3 / 65.6               & 59.3 / 61.7               & 78.4 / \underline{76.9} \\
2020 $\rightarrow$ 2021-extra                       
& 64.5 / 65.5               & 59.5 / 61.4               & 78.6 / 76.3 \\
\bottomrule
\end{tabular}
}
\caption{
Results of the self-labeling experiment with different strategies for RoBERTa\textsubscript{LARGE} model ($\rightarrow$: continuous fine-tuning; +: concatenation of datasets) where micro and macro F1 score as well as type-ignored F1 score on the test set of 2021-set / 2020-set are reported.
The best results in each of the 2021-set / 2020-set are highlighted in bold character / underline in each metric.}
\label{tab:self-labeling}
\end{table}

\subsubsection{Evaluation}

\noindent\textbf{Experimental Setting.} For our experiments we focused on the best model in our previous experiments, which is RoBERTa\textsubscript{LARGE}.
We collected extra (unlabeled) tweets following the same procedure described in \cite{dimosthenis-etal-2022-twitter}, that results in 93,594 and 878,80 tweets from the period of 2020-set and 2021-set, respectively. Over those extra tweets, we use the RoBERTa\textsubscript{LARGE} NER model fine-tuned on the 2020-set to predict labels.

\noindent\textbf{Results.} \autoref{tab:self-labeling} shows the result of self-labeling, where we report three patterns of model fine-tuning: (i) fine-tuning only on the pseudo dataset (e.g., 2020-extra); (ii) fine-tuning on the joint dataset of the training set of the 2020-set and the pseudo dataset (e.g., 2020 + 2020-extra); and (iii) continuous fine-tuning of the 2020-set fine-tuned model on the pseudo dataset (e.g., 2020 $\rightarrow$ 2020-extra).
In general, we can not find any major improvement by self-labeling, regardless of the strategy. In a way, this contradicts the self-labeling experiment on the TTC dataset performed by \newcite{agarwal2021temporal}.\footnote{While in our setting we extract a larger number of tweets, this trend does not change with less self-labeled training data.} This may suggest that the temporal-shift of TweetNER7 is more challenging to mitigate than TTC, and self-labeling is not enough in itself to overcome the temporal shift.

\subsubsection{Contextual Prediction Analysis}

To explore the reason why self-labeling does not help to mitigate temporal-shift in TweetNER7, we conducted an analysis over the self-labeled tweets.
Inspired by recent semi-parametric approach in information retrieval \cite{lewis-etal-2021-paq}, we considered a retrieval module that fetches relevant tweets given a target entity from the self-labeled corpus and see the portion of retrieved tweets containing the true prediction. To be precise, we first ran the NER model prediction on target tweets, and for each of the predicted entities. Then, we queried tweets from the extra tweet corpus used in Subsection \ref{sec:self-labeling} to compute the ratio of correct predictions within the retrieved predictions, which we call contextualized predictions.
Since we are interested in the error of the original prediction, we focus only on the entities where the original prediction is incorrect.

\begin{figure}[t]
    \centering
    \includegraphics[width=1.0\columnwidth]{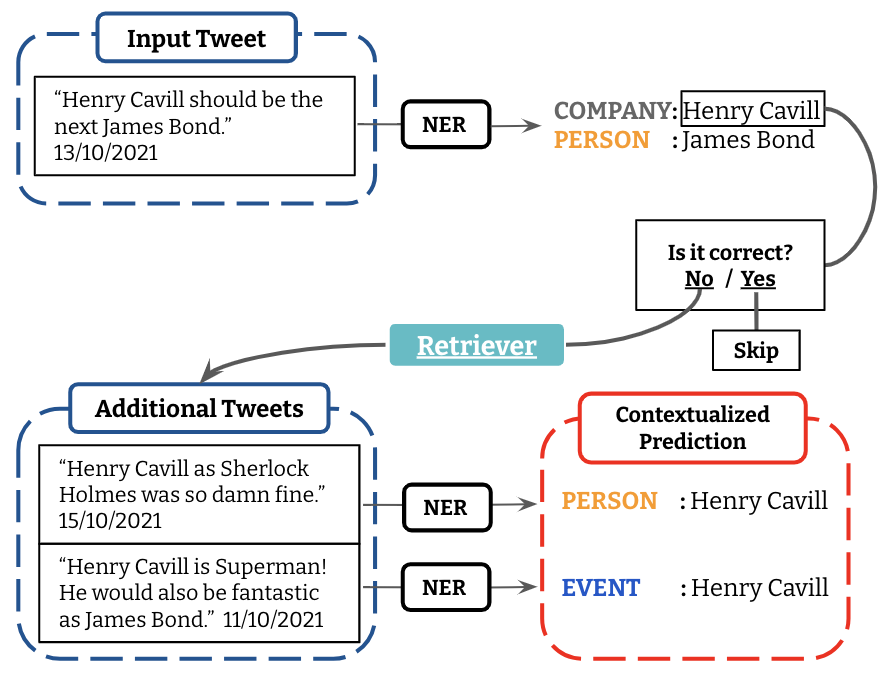}
    \caption{Overview of the pipeline to retrieve contextualized prediction.
    }
    \label{fig:contextual-prediction}
\end{figure}

\autoref{fig:contextual-prediction} describes the whole pipeline and we use Whoosh library\footnote{\url{https://pypi.org/project/Whoosh/}} for search engine where the query is always the entity name, constraining the search result by the number of days from the query tweet.\footnote{Setting days as 7 means the search results should be in the range of 7 days before/after was made.}
Similarly to the analysis in \autoref{sec:self-labeling}, we used the RoBERTa\textsubscript{LARGE} fine-tuned on the 2020-set of TweetNER7 and evaluated the contextualized predictions on the 2021 test set.

\begin{figure}[t]
    \centering
    \includegraphics[width=1.0\columnwidth]{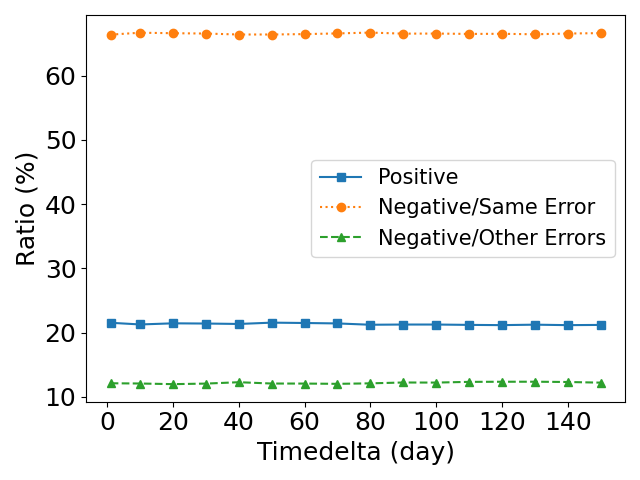}
    \caption{
    Ratio of positive and negative predictions in the contextualized tweets, split into two error types: same prediction as the original prediction or not. The X-axis represents the days from the original tweet (0=same date as the original tweet) and results are broken on 20-day chunks..
    }
    \label{fig:contextual-result}
\end{figure}

\autoref{fig:contextual-result} shows the ratio of positive and negative predictions in the contextualized tweets. These are further broken into two error types whether it is the same prediction as the original prediction or not, along with the days we set as a search constraint.
Most frequent predictions are usually the same as the original predictions, which means that the original language model tends to output similar predictions for the same entities, irrespective of the context. As far as the time variable is concerned, the ratio is almost consistent over time, which suggests that the possible original bias of the model does not change over time. Nonetheless, the second most frequent predictions are on average the correct ones, with a large gap with respect to other types of error. This implies there may still be a useful signal to improve the original prediction in the self-labeled corpus.

\section{Conclusion}
In this paper, we have constructed TweetNER7, a new NER dataset for Twitter, in which we annotated 11,382 tweets with seven entity types. The collected tweets are distributed uniformly over time from September 2019 to August 2021, which facilitates temporal analysis in NER for social media.
The dataset is diverse topic-wise, as we leveraged weekly trending topics to query tweets and near-duplicated and irrelevant tweets were dropped.
To establish baselines on TweetNER7, we fine-tuned standard LMs including a few Twitter-specific LMs. Moreover, we performed a few targeted temporal-related analyses in order to better understand the short-term temporal effect.
Finally, we show that self-labeling is not enough to mitigate the temporal-shift and had no noticeable improvement over the baseline vanilla fine-tuning, which further highlights the challenging nature of the dataset.

\section{Limitations and Future Work}
The TweetNER7 dataset was constructed on English tweets so it is limited to English, as most of the existing NER datasets for social media \cite{derczynski-etal-2016-broad}. In the future we are planning to apply a similar methodology to extend it to languages other than English. Given the dynamic nature of social media, TweetNER7 is designed to study short-term temporal-shift (e.g., monthly) but would not be suitable for analysing longer temporal shifts (e.g., yearly) \cite{rijhwani-preotiuc-pietro-2020-temporally}. 
We selected Twitter as the data source but temporal-shift is a common problem in social media generally. As a future work, we are planning to add more data from other social media platforms as in WNUT17 \citep{derczynski-etal-2017-results} to give us more general insights to understand temporal shift phenomena in social media more generally.

\section*{Acknowledgements}

Jose Camacho-Collados is supported by a UKRI Future Leaders Fellowship.





\bibliography{anthology,custom}

\begin{thebibliography}{32}
\expandafter\ifx\csname natexlab\endcsname\relax\def\natexlab#1{#1}\fi

\bibitem[{Agarwal and Nenkova(2022)}]{agarwal2021temporal}
Oshin Agarwal and Ani Nenkova. 2022.
\newblock \href {https://doi.org/10.1162/tacl_a_00497} {Temporal effects on
  pre-trained models for language processing tasks}.
\newblock \emph{Transactions of the Association for Computational Linguistics},
  10:904--921.

\bibitem[{Antypas et~al.(2022)Antypas, Ushio, Camacho-Collados, Neves, Silva,
  and Barbieri}]{dimosthenis-etal-2022-twitter}
Dimosthenis Antypas, Asahi Ushio, Jose Camacho-Collados, Leonardo Neves, Vitor
  Silva, and Francesco Barbieri. 2022.
\newblock {T}witter {T}opic {C}lassification.
\newblock In \emph{Proceedings of the 29th International Conference on
  Computational Linguistics}, Gyeongju, Republic of Korea. International
  Committee on Computational Linguistics.

\bibitem[{Camacho-Collados et~al.(2022)Camacho-Collados, Rezaee, Riahi, Ushio,
  Loureiro, Antypas, Boisson, Espinosa-Anke, Liu, Mart{\'\i}nez-C{\'a}mara
  et~al.}]{camacho2022tweetnlp}
Jose Camacho-Collados, Kiamehr Rezaee, Talayeh Riahi, Asahi Ushio, Daniel
  Loureiro, Dimosthenis Antypas, Joanne Boisson, Luis Espinosa-Anke, Fangyu
  Liu, Eugenio Mart{\'\i}nez-C{\'a}mara, et~al. 2022.
\newblock Tweetnlp: Cutting-edge natural language processing for social media.
\newblock \emph{arXiv preprint arXiv:2206.14774}.

\bibitem[{Collier and Kim(2004)}]{collier-kim-2004-introduction}
Nigel Collier and Jin-Dong Kim. 2004.
\newblock \href {https://aclanthology.org/W04-1213} {Introduction to the
  bio-entity recognition task at {JNLPBA}}.
\newblock In \emph{Proceedings of the International Joint Workshop on Natural
  Language Processing in Biomedicine and its Applications
  ({NLPBA}/{B}io{NLP})}, pages 73--78, Geneva, Switzerland. COLING.

\bibitem[{Del~Tredici et~al.(2019)Del~Tredici, Fern{\'a}ndez, and
  Boleda}]{del-tredici-etal-2019-short}
Marco Del~Tredici, Raquel Fern{\'a}ndez, and Gemma Boleda. 2019.
\newblock \href {https://doi.org/10.18653/v1/N19-1210} {Short-term meaning
  shift: A distributional exploration}.
\newblock In \emph{Proceedings of the 2019 Conference of the North {A}merican
  Chapter of the Association for Computational Linguistics: Human Language
  Technologies, Volume 1 (Long and Short Papers)}, pages 2069--2075,
  Minneapolis, Minnesota. Association for Computational Linguistics.

\bibitem[{Derczynski et~al.(2016)Derczynski, Bontcheva, and
  Roberts}]{derczynski-etal-2016-broad}
Leon Derczynski, Kalina Bontcheva, and Ian Roberts. 2016.
\newblock \href {https://aclanthology.org/C16-1111} {Broad {T}witter corpus: A
  diverse named entity recognition resource}.
\newblock In \emph{Proceedings of {COLING} 2016, the 26th International
  Conference on Computational Linguistics: Technical Papers}, pages 1169--1179,
  Osaka, Japan. The COLING 2016 Organizing Committee.

\bibitem[{Derczynski et~al.(2017)Derczynski, Nichols, van Erp, and
  Limsopatham}]{derczynski-etal-2017-results}
Leon Derczynski, Eric Nichols, Marieke van Erp, and Nut Limsopatham. 2017.
\newblock \href {https://doi.org/10.18653/v1/W17-4418} {Results of the
  {WNUT}2017 shared task on novel and emerging entity recognition}.
\newblock In \emph{Proceedings of the 3rd Workshop on Noisy User-generated
  Text}, pages 140--147, Copenhagen, Denmark. Association for Computational
  Linguistics.

\bibitem[{Devlin et~al.(2019)Devlin, Chang, Lee, and
  Toutanova}]{devlin-etal-2019-bert}
Jacob Devlin, Ming-Wei Chang, Kenton Lee, and Kristina Toutanova. 2019.
\newblock \href {https://doi.org/10.18653/v1/N19-1423} {{BERT}: Pre-training of
  deep bidirectional transformers for language understanding}.
\newblock In \emph{Proceedings of the 2019 Conference of the North {A}merican
  Chapter of the Association for Computational Linguistics: Human Language
  Technologies, Volume 1 (Long and Short Papers)}, pages 4171--4186,
  Minneapolis, Minnesota. Association for Computational Linguistics.

\bibitem[{Hovy et~al.(2006)Hovy, Marcus, Palmer, Ramshaw, and
  Weischedel}]{hovy-etal-2006-ontonotes}
Eduard Hovy, Mitchell Marcus, Martha Palmer, Lance Ramshaw, and Ralph
  Weischedel. 2006.
\newblock \href {https://aclanthology.org/N06-2015} {{O}nto{N}otes: The 90{\%}
  solution}.
\newblock In \emph{Proceedings of the Human Language Technology Conference of
  the {NAACL}, Companion Volume: Short Papers}, pages 57--60, New York City,
  USA. Association for Computational Linguistics.

\bibitem[{Howard and Ruder(2018)}]{howard-ruder-2018-universal}
Jeremy Howard and Sebastian Ruder. 2018.
\newblock \href {https://doi.org/10.18653/v1/P18-1031} {Universal language
  model fine-tuning for text classification}.
\newblock In \emph{Proceedings of the 56th Annual Meeting of the Association
  for Computational Linguistics (Volume 1: Long Papers)}, pages 328--339,
  Melbourne, Australia. Association for Computational Linguistics.

\bibitem[{Jiang et~al.(2022)Jiang, Hua, Beeferman, and
  Roy}]{jiang-EtAl:2022:LREC2}
Hang Jiang, Yining Hua, Doug Beeferman, and Deb Roy. 2022.
\newblock \href {https://aclanthology.org/2022.lrec-1.780} {Annotating the
  tweebank corpus on named entity recognition and building nlp models for
  social media analysis}.
\newblock In \emph{Proceedings of the Language Resources and Evaluation
  Conference}, pages 7199--7208, Marseille, France. European Language Resources
  Association.

\bibitem[{Lazaridou et~al.(2021)Lazaridou, Kuncoro, Gribovskaya, Agrawal,
  Liska, Terzi, Gimenez, de~Masson~d'Autume, Kocisky, Ruder
  et~al.}]{lazaridou2021mind}
Angeliki Lazaridou, Adhi Kuncoro, Elena Gribovskaya, Devang Agrawal, Adam
  Liska, Tayfun Terzi, Mai Gimenez, Cyprien de~Masson~d'Autume, Tomas Kocisky,
  Sebastian Ruder, et~al. 2021.
\newblock Mind the gap: Assessing temporal generalization in neural language
  models.
\newblock \emph{Advances in Neural Information Processing Systems},
  34:29348--29363.

\bibitem[{Lee et~al.(2020)Lee, Yoon, Kim, Kim, Kim, So, and
  Kang}]{lee2020biobert}
Jinhyuk Lee, Wonjin Yoon, Sungdong Kim, Donghyeon Kim, Sunkyu Kim, Chan~Ho So,
  and Jaewoo Kang. 2020.
\newblock Biobert: a pre-trained biomedical language representation model for
  biomedical text mining.
\newblock \emph{Bioinformatics}, 36(4):1234--1240.

\bibitem[{Lewis et~al.(2021)Lewis, Wu, Liu, Minervini, K{\"u}ttler, Piktus,
  Stenetorp, and Riedel}]{lewis-etal-2021-paq}
Patrick Lewis, Yuxiang Wu, Linqing Liu, Pasquale Minervini, Heinrich
  K{\"u}ttler, Aleksandra Piktus, Pontus Stenetorp, and Sebastian Riedel. 2021.
\newblock \href {https://doi.org/10.1162/tacl_a_00415} {{PAQ}: 65 million
  probably-asked questions and what you can do with them}.
\newblock \emph{Transactions of the Association for Computational Linguistics},
  9:1098--1115.

\bibitem[{Li et~al.(2016)Li, Sun, Johnson, Sciaky, Wei, Leaman, Davis,
  Mattingly, Wiegers, and Lu}]{li2016biocreative}
Jiao Li, Yueping Sun, Robin~J Johnson, Daniela Sciaky, Chih-Hsuan Wei, Robert
  Leaman, Allan~Peter Davis, Carolyn~J Mattingly, Thomas~C Wiegers, and Zhiyong
  Lu. 2016.
\newblock Biocreative v cdr task corpus: a resource for chemical disease
  relation extraction.
\newblock \emph{Database}, 2016.

\bibitem[{Li et~al.(2020)Li, Sun, Meng, Liang, Wu, and Li}]{li-etal-2020-dice}
Xiaoya Li, Xiaofei Sun, Yuxian Meng, Junjun Liang, Fei Wu, and Jiwei Li. 2020.
\newblock \href {https://doi.org/10.18653/v1/2020.acl-main.45} {Dice loss for
  data-imbalanced {NLP} tasks}.
\newblock In \emph{Proceedings of the 58th Annual Meeting of the Association
  for Computational Linguistics}, pages 465--476, Online. Association for
  Computational Linguistics.

\bibitem[{Liu et~al.(2018)Liu, Zhu, Che, Qin, Schneider, and
  Smith}]{liu-etal-2018-parsing}
Yijia Liu, Yi~Zhu, Wanxiang Che, Bing Qin, Nathan Schneider, and Noah~A. Smith.
  2018.
\newblock \href {https://doi.org/10.18653/v1/N18-1088} {Parsing tweets into
  {U}niversal {D}ependencies}.
\newblock In \emph{Proceedings of the 2018 Conference of the North {A}merican
  Chapter of the Association for Computational Linguistics: Human Language
  Technologies, Volume 1 (Long Papers)}, pages 965--975, New Orleans,
  Louisiana. Association for Computational Linguistics.

\bibitem[{Liu et~al.(2019)Liu, Ott, Goyal, Du, Joshi, Chen, Levy, Lewis,
  Zettlemoyer, and Stoyanov}]{RoBERTa}
Yinhan Liu, Myle Ott, Naman Goyal, Jingfei Du, Mandar Joshi, Danqi Chen, Omer
  Levy, Mike Lewis, Luke Zettlemoyer, and Veselin Stoyanov. 2019.
\newblock \href {http://arxiv.org/abs/1907.11692} {{RoBERTa}: {A} robustly
  optimized {BERT} pretraining approach}.
\newblock \emph{CoRR}, abs/1907.11692.

\bibitem[{Loureiro et~al.(2022)Loureiro, Barbieri, Neves, Espinosa~Anke, and
  Camacho-collados}]{loureiro-etal-2022-timelms}
Daniel Loureiro, Francesco Barbieri, Leonardo Neves, Luis Espinosa~Anke, and
  Jose Camacho-collados. 2022.
\newblock \href {https://doi.org/10.18653/v1/2022.acl-demo.25} {{T}ime{LM}s:
  Diachronic language models from {T}witter}.
\newblock In \emph{Proceedings of the 60th Annual Meeting of the Association
  for Computational Linguistics: System Demonstrations}, pages 251--260,
  Dublin, Ireland. Association for Computational Linguistics.

\bibitem[{Nguyen et~al.(2020)Nguyen, Vu, and
  Tuan~Nguyen}]{nguyen-etal-2020-bertweet}
Dat~Quoc Nguyen, Thanh Vu, and Anh Tuan~Nguyen. 2020.
\newblock \href {https://doi.org/10.18653/v1/2020.emnlp-demos.2} {{BERT}weet: A
  pre-trained language model for {E}nglish tweets}.
\newblock In \emph{Proceedings of the 2020 Conference on Empirical Methods in
  Natural Language Processing: System Demonstrations}, pages 9--14, Online.
  Association for Computational Linguistics.

\bibitem[{Pan et~al.(2017)Pan, Zhang, May, Nothman, Knight, and
  Ji}]{pan-etal-2017-cross}
Xiaoman Pan, Boliang Zhang, Jonathan May, Joel Nothman, Kevin Knight, and Heng
  Ji. 2017.
\newblock \href {https://doi.org/10.18653/v1/P17-1178} {Cross-lingual name
  tagging and linking for 282 languages}.
\newblock In \emph{Proceedings of the 55th Annual Meeting of the Association
  for Computational Linguistics (Volume 1: Long Papers)}, pages 1946--1958,
  Vancouver, Canada. Association for Computational Linguistics.

\bibitem[{Peters et~al.(2018)Peters, Neumann, Iyyer, Gardner, Clark, Lee, and
  Zettlemoyer}]{peters-etal-2018-deep}
Matthew~E. Peters, Mark Neumann, Mohit Iyyer, Matt Gardner, Christopher Clark,
  Kenton Lee, and Luke Zettlemoyer. 2018.
\newblock \href {https://doi.org/10.18653/v1/N18-1202} {Deep contextualized
  word representations}.
\newblock In \emph{Proceedings of the 2018 Conference of the North {A}merican
  Chapter of the Association for Computational Linguistics: Human Language
  Technologies, Volume 1 (Long Papers)}, pages 2227--2237, New Orleans,
  Louisiana. Association for Computational Linguistics.

\bibitem[{Radford et~al.(2018)Radford, Narasimhan, Salimans, and
  Sutskever}]{GPT}
Alec Radford, Karthik Narasimhan, Tim Salimans, and Ilya Sutskever. 2018.
\newblock Improving language understanding by generative pre-training.
\newblock \emph{Technical report, OpenAI}.

\bibitem[{Radford et~al.(2019)Radford, Wu, Child, Luan, Amodei, and
  Sutskever}]{GPT2}
Alec Radford, Jeffrey Wu, Rewon Child, David Luan, Dario Amodei, and Ilya
  Sutskever. 2019.
\newblock Language models are unsupervised multitask learners.
\newblock \emph{OpenAI blog}, 1(8):9.

\bibitem[{Rijhwani and
  Preotiuc-Pietro(2020)}]{rijhwani-preotiuc-pietro-2020-temporally}
Shruti Rijhwani and Daniel Preotiuc-Pietro. 2020.
\newblock \href {https://doi.org/10.18653/v1/2020.acl-main.680}
  {Temporally-informed analysis of named entity recognition}.
\newblock In \emph{Proceedings of the 58th Annual Meeting of the Association
  for Computational Linguistics}, pages 7605--7617, Online. Association for
  Computational Linguistics.

\bibitem[{Salinas~Alvarado et~al.(2015)Salinas~Alvarado, Verspoor, and
  Baldwin}]{salinas-alvarado-etal-2015-domain}
Julio~Cesar Salinas~Alvarado, Karin Verspoor, and Timothy Baldwin. 2015.
\newblock \href {https://aclanthology.org/U15-1010} {Domain adaption of named
  entity recognition to support credit risk assessment}.
\newblock In \emph{Proceedings of the Australasian Language Technology
  Association Workshop 2015}, pages 84--90, Parramatta, Australia.

\bibitem[{Tedeschi and Navigli(2022)}]{tedeschi-navigli-2022-multinerd}
Simone Tedeschi and Roberto Navigli. 2022.
\newblock \href {https://aclanthology.org/2022.findings-naacl.60}
  {{M}ulti{NERD}: A multilingual, multi-genre and fine-grained dataset for
  named entity recognition (and disambiguation)}.
\newblock In \emph{Findings of the Association for Computational Linguistics:
  NAACL 2022}, pages 801--812, Seattle, United States. Association for
  Computational Linguistics.

\bibitem[{Tjong Kim~Sang and
  De~Meulder(2003)}]{tjong-kim-sang-de-meulder-2003-introduction}
Erik~F. Tjong Kim~Sang and Fien De~Meulder. 2003.
\newblock \href {https://aclanthology.org/W03-0419} {Introduction to the
  {C}o{NLL}-2003 shared task: Language-independent named entity recognition}.
\newblock In \emph{Proceedings of the Seventh Conference on Natural Language
  Learning at {HLT}-{NAACL} 2003}, pages 142--147.

\bibitem[{Ushio and Camacho-Collados(2021)}]{ushio-camacho-collados-2021-ner}
Asahi Ushio and Jose Camacho-Collados. 2021.
\newblock \href {https://doi.org/10.18653/v1/2021.eacl-demos.7} {{T}-{NER}: An
  all-round python library for transformer-based named entity recognition}.
\newblock In \emph{Proceedings of the 16th Conference of the European Chapter
  of the Association for Computational Linguistics: System Demonstrations},
  pages 53--62, Online. Association for Computational Linguistics.

\bibitem[{Wei et~al.(2015)Wei, Peng, Leaman, Davis, Mattingly, Li, Wiegers, and
  Lu}]{wei2015overview}
Chih-Hsuan Wei, Yifan Peng, Robert Leaman, Allan~Peter Davis, Carolyn~J
  Mattingly, Jiao Li, Thomas~C Wiegers, and Zhiyong Lu. 2015.
\newblock Overview of the biocreative v chemical disease relation (cdr) task.
\newblock In \emph{Proceedings of the fifth BioCreative challenge evaluation
  workshop}, volume~14.

\bibitem[{Wolf et~al.(2020)Wolf, Debut, Sanh, Chaumond, Delangue, Moi, Cistac,
  Rault, Louf, Funtowicz, Davison, Shleifer, von Platen, Ma, Jernite, Plu, Xu,
  Le~Scao, Gugger, Drame, Lhoest, and Rush}]{wolf-etal-2020-transformers}
Thomas Wolf, Lysandre Debut, Victor Sanh, Julien Chaumond, Clement Delangue,
  Anthony Moi, Pierric Cistac, Tim Rault, Remi Louf, Morgan Funtowicz, Joe
  Davison, Sam Shleifer, Patrick von Platen, Clara Ma, Yacine Jernite, Julien
  Plu, Canwen Xu, Teven Le~Scao, Sylvain Gugger, Mariama Drame, Quentin Lhoest,
  and Alexander Rush. 2020.
\newblock \href {https://doi.org/10.18653/v1/2020.emnlp-demos.6} {Transformers:
  State-of-the-art natural language processing}.
\newblock In \emph{Proceedings of the 2020 Conference on Empirical Methods in
  Natural Language Processing: System Demonstrations}, pages 38--45, Online.
  Association for Computational Linguistics.

\bibitem[{Yamada et~al.(2020)Yamada, Asai, Shindo, Takeda, and
  Matsumoto}]{yamada-etal-2020-luke}
Ikuya Yamada, Akari Asai, Hiroyuki Shindo, Hideaki Takeda, and Yuji Matsumoto.
  2020.
\newblock \href {https://doi.org/10.18653/v1/2020.emnlp-main.523} {{LUKE}: Deep
  contextualized entity representations with entity-aware self-attention}.
\newblock In \emph{Proceedings of the 2020 Conference on Empirical Methods in
  Natural Language Processing (EMNLP)}, pages 6442--6454, Online. Association
  for Computational Linguistics.

\end{thebibliography}
\bibliographystyle{acl_natbib}

\end{document}